\documentclass[10pt,twocolumn,letterpaper]{article}

\usepackage{cvpr}
\usepackage{times}
\usepackage{epsfig}
\usepackage{graphicx}
\usepackage{amsmath}
\usepackage{amssymb}

\usepackage{algorithm}
\usepackage{algorithmic}
\usepackage{appendix}
\usepackage{subcaption}
\usepackage{graphicx}

\usepackage[utf8]{inputenc} 
\usepackage[T1]{fontenc}    
\usepackage{url}            
\usepackage{amsfonts}       
\usepackage{nicefrac}       

\usepackage{mathtools}

\usepackage{microtype}
\usepackage{booktabs}\usepackage{color}
\usepackage{cases}
\usepackage{sidecap}
\usepackage{wrapfig}
\usepackage{leftidx}
\usepackage{pifont}
\usepackage[english]{babel}
\usepackage{enumitem}
\usepackage{bm}
\usepackage{amsthm}
\usepackage{multirow}
\usepackage{colortbl}
\usepackage{mathrsfs}

\usepackage[pagebackref=true,breaklinks=true,letterpaper=true,colorlinks,bookmarks=false]{hyperref}

\cvprfinalcopy 

\ifcvprfinal\fi

\newcommand{\eref}[1]{Eq.~(\ref{#1})}

\newcommand{\bx}{\bm{x}}
\newcommand{\bz}{\bm{z}}

\newcommand{\pg}{p_g}
\newcommand{\pgr}{p_{gr}}
\newcommand{\pgf}{p_{gf}}
\newcommand{\pd}{p_{data}}
\newcommand{\pz}{p_z}

\newcommand{\xg}{\bx_g}
\newcommand{\xgr}{\bx_{gr}}
\newcommand{\xgf}{\bx_{gf}}
\newcommand{\xd}{\bx_{data}}

\newtheorem{theorem}{Theorem}

\newtheorem{corollary}{Corollary}

\begin{document}

\title{On Positive-Unlabeled Classification in GAN}

\author{Tianyu Guo$^{1,2}$, Chang Xu$^{2}$, Jiajun Huang$^{2}$, Yunhe Wang$^3$, Boxin Shi$^{4,5}$, Chao Xu$^1$, Dacheng Tao$^2$\\
  \small$^1$Key Laboratory of Machine Perception (MOE), CMIC,
School of EECS, Peking University, China\\
\small$^2$UBTECH Sydney AI Centre, School of Computer Science, Faculty of Engineering, The University of Sydney, Australia\\
\small$^3$ Huawei Noah's Ark Lab 
\small$^4$National Engineering Laboratory for Video Technology, Peking University $^5$Peng Cheng Laboratory\\
  \small\texttt{\{tianyuguo, shiboxin\}@pku.edu.cn}, \texttt{chaoxu@cis.pku.edu.cn, yunhe.wang@huawei.com}\\
  \small\texttt{\{c.xu, dacheng.tao\}@sydney.edu.au}, \texttt{jhua7177@uni.sydney.edu.au} \\
}

\maketitle

\begin{abstract}
This paper defines a positive and unlabeled classification problem for standard GANs, which then leads to a novel technique to stabilize the training of the discriminator in GANs. Traditionally, real data are taken as positive while generated data are negative. This positive-negative classification criterion was kept fixed all through the learning process of the discriminator without considering the gradually improved quality of generated data, even if they could be more realistic than real data at times. In contrast, it is more reasonable to treat the generated data as unlabeled, which could be positive or negative according to their quality. The discriminator is thus a classifier for this positive and unlabeled classification problem, and we derive a new Positive-Unlabeled GAN (PUGAN). We theoretically discuss the global optimality the proposed model will achieve and the equivalent optimization goal. Empirically, we find that PUGAN can achieve comparable or even better performance than those sophisticated discriminator stabilization methods.


\end{abstract}

\section{Introduction}
Recently, deep generative models have received remarkable achievements in image generation tasks~\cite{kingma2013auto,oord2016pixel,van2016conditional,goodfellow2014generative}. As a representative generative model, GANs~\cite{goodfellow2014generative} approximated a target distribution via playing a min-max game. In the standard framework of GAN~\cite{goodfellow2014generative,radford2015unsupervised}, a generator takes noise vectors from a prior distribution (\eg Gaussian distribution and normal distribution) as the input and tends to produce data that follows the distribution of the reference natural images, while the discriminator aims to distinguish the generated data from the real data. Various GAN methods have been developed in many interesting applications. For example, in the image-to-image translation task, generators in GANs map the input image to output image. Representative methods include Pix2pix~\cite{isola2017image} over paired training images and cycleGAN~\cite{zhu2017unpaired} in an unsupervised way. 

In vanilla GANs, the training progress usually lacks stability, and the quality of generated images is not always satisfactory (\eg model collapse).  For instance, DCGAN~\cite{radford2015unsupervised} carefully designed the neural architectures for the generator and the discriminator to alleviate these problems. Progressive GAN~\cite{karras2017progressive} generated high-resolution images by progressively deepening the network. BigGAN~\cite{brock2018large} produced high-quality images by improving training methods, \eg enlarging batch size, and truncating the latent space.
WGAN~\cite{arjovsky2017wasserstein} and WGAN-GP~\cite{gulrajani2017improved} tried to fit and optimize the Wasserstein distance to stabilize the generation process. SNGAN~\cite{miyato2018spectral} proved the necessity and benefits of introducing Lipschitz continuity to the discriminator.

These aforementioned methods to stabilize GAN could be roughly divided into two categories: designing stable network structures and training strategies and developing new effective optimization goals. However, neither of them has stepped away from the positive-negative classification problem initially established in standard GAN. Although WGAN and WGAN-GP no longer take the discriminator as a classifier for real data and generated data, the aim of the is still to separate the real and generated data as far as possible. To the best of our knowledge, existing GAN models attempt to strictly distinguish between generated data and real data and ignore the fact that the quality of the generated samples is not the same. It is unfair to treat high-quality samples equally with low-quality samples, especially when high-quality samples are sufficiently realistic. Although there are many theoretical results proposed to justify the final equilibrium, such as vanilla GAN~\cite{goodfellow2014generative} proving the existing of the equilibrium and WGAN~\cite{arjovsky2017wasserstein} replacing the JS divergence with the Wasserstein distance, these analysis mainly focus on the final achievement rather than the intermediate status in the training process.

In this paper, we suggest that instead of an ordinary positive and negative classification (\ie real v.s. fake) problem, GAN is actually in the face of a positive-unlabeled classification problem. With adequate training, generated data could look real and may appear to be even more realistic than real data at times. It would then be illogical to make a stereotype of generated data as fake data. To catch up with the continuously improved quality of generated data, we thus take them as unlabeled data, which consists of low-quality data and high-quality data. These high-quality data are considered to be close to or even better than some real data. Within the framework of positive-unlabeled classification,  the classification objective of standard GAN can be re-defined, and different variants can be easily obtained by considering different scoring functions (\eg those in LSGAN \cite{LSGAN} and HingeGAN \cite{miyato2018spectral}). In addition, we get rid of the class balance constraint (\ie half of the sample are fake), and observe impressive performance improvement by increasing the share of generated data in the mini-batch. Our theoretical analysis suggests that the proposed new algorithm has a guaranteed final equilibrium. Experimental results on benchmark datasets demonstrate that we actually enjoy more stable training progress and thus achieve better generated samples. 



\section{Proposed Approach}\label{sec_3}


In this section, we first review preliminary works about the standard Generative Adversarial Network (SGAN). Then we analyze the problem existing in GAN and define a new role for the discriminator $D$. We also theoretically develop this idea into a new algorithm within the framework of SGAN, and then extended this algorithm to the general GAN, which shows the flexibility of our method.

\subsection{Preliminary}

GAN~\cite{goodfellow2014generative} was introduced by Goodfellow et al. (2014).  It consists of two neural networks: discriminator network $D$ and generator network $G$. The discriminator $D$ aims to distinguish the provided real data and the fake data generated by the generator $G$. On the other hand, the generator $G$ aims to generate fake data that can fool the discriminator $D$. Following this adversarial manner, we expect the $G$ can generate high-quality data in the end. Formally, the objective function of GAN  can be written as
\begin{equation}
\begin{split}
	\underset{G}{\min}~\underset{D}{\max} ~ V(D,G) &= \mathbb{E}_{\bx\sim p_{data}(\bx)}[\log D(\bx)] \\ 
	&+ \mathbb{E}_{\bz\sim p_{\bz}(\bz)}[\log (1-D(G(\bz)))],
\end{split}    
\end{equation}
where \(p_{data}\) indicates the distribution of real data, $\bz$ is the random noise sampled from a prior distribution \(p_{\bz}\) (\ie the Gaussian distribution), and $D(\bx)$ is the predicted probability of $\bx$ to be real by the discriminator. Since the minimax objective function might lead to gradient vanishing for $G$ when $D$ can perfectly distinguish two data set. More of GAN's variants (\eg WGAN~\cite{arjovsky2017wasserstein} and LSGAN~\cite{mao2017least}) transform this minimax game into a non-saturating game. In general, the objective functions of these GANs can be concluded as follows:
\begin{equation}\label{general_d}
\begin{split}
	 \underset{G}{\min}~\underset{D}{\max}~ V(D, G) &= \mathbb{E}_{\bx\sim p_{data}(\bx)}[f_{1} (D(\bx))] \\
	&- \mathbb{E}_{\bz\sim p_{\bz}(\bz)}[f_{2}(D(G(\bz)))],
\end{split}
\end{equation}
where \(f_1\) is the loss of classifying input as real and \(f_2\) is the loss of classifying input as fake.

\subsection{Problem Analysis}

%


As shown above, existing GAN variants are trained to separate the real and generated data strictly. However, this does not match the actual situation in training. Some of the generated samples can achieve higher quality and are more realistic than others. This phenomenon usually lasts until the end of the training. As a result, the quality of samples generated by $G$ is very different, and there are many high-quality samples and a considerable proportion of low-quality samples. For instance, as a well-known problem in GAN, the model collapse problem that $G$ networks often suffer can be considered as the generation space consisting of some high quality and non-repetitive samples and the rest of the repetitive samples. These duplicate samples can be considered as low-quality samples and still need to be improved. There is still a phenomenon that there is a certain proportion of unsatisfactory samples in a well-trained generation space, and the gaps in the generated samples of different quality are relatively large. Therefore, the traditional method of strictly distinguishing the real sample from the generated sample does not conform to the actual situation of the training. In this paper, we propose an algorithm that is dedicated to picking out low-quality samples from the samples generated by $G$ and promoting them, unlike traditional discriminators $D$ that are dedicated to distinguishing real samples from generated samples. Our algorithm encourages the discriminator $D$ to divide the generation space of $G$ into high-quality samples and low-quality samples so that the generator $G$ could improve the low-quality samples.


By doing so, the proposed method enjoys several desired properties: i) The discriminator pays more attention to poor quality samples, allowing the generator to focus on improving the quality of these bad samples. As a result, the quality of generated samples is more balanced, and the overall quality is expected to be enhanced, ii) The training strategy of our algorithm is more in line with the actual situation of samples generated by $G$ in the training process, so the training process is expected to be more stable, iii) More importantly, the proposed algorithm is a flexible method which means that our algorithm could be easily integrated into existing frameworks of variety GAN, and we will show this desirable feature in the following subsection, and iv) Although the proposed method changes the function of the D network, we provide some theoretical results in Section~\ref{theoretical_results} which demonstrate the proposed method also enjoys the same equilibrium condition with the other GAN and provide some guarantee for performance.


\subsection{Positive-Unlabeled classification (PU) in SGAN}
Above, we discuss the current problem existing in the GAN framework and propose to allow some good samples to be recognized as real data. In this part, we firstly introduce how we achieve this in the framework of standard GAN, and then we extend it for general GANs.

%

As mentioned above, we propose to allow the discriminator $D$ to treat the high-quality generated samples as real data and focus on the bad generated samples. The discriminator $D$ is required to learn how to distinguish high-quality samples with other low-quality samples. Identifying high-quality samples from generated samples under the guidance of real samples is very similar to Positive-Unlabeled classification problems~\cite{du2014analysis,kiryo2017positive,xu2017multi}, where only some positive samples were labeled, and the classifier tried to find positive samples from unlabeled samples consisting of positive and negative samples. According to the solution of the PU classification problem, we develop an algorithm learning a discriminator to recognize high-quality samples. Firstly, we denote the generated data as $\xg$, consisted with high-quality samples $\xgr$ and the bad samples $\xgf$. And we consider both $\xgr$ and real data $\xd$ to be real (\ie $y_{gr} = y_{data} = 1$) while consider $\xgf$ to be fake (\ie $y_{gf} = -1$). In addition, denote $p(x)$ as the marginal density of $\xg$ and $\pgr (\bx) = p(\bx|1)$ and $p_{gf}(\bx) = p(\bx|-1)$ are the class conditional densities of $\bx_{gr}$ and $\bx_{gf}$ respectively. Then the $p_g$ which is the marginal density of $x_g$, can be obtained with:
\begin{equation}\label{eq_pi}
    p_g(\bx) = \pi p_{gr}(\bx) + (1-\pi) p_{gf}(\bx),
\end{equation}
where $\pi$ is the unknown \textit{class prior} (\ie, the proportion of $\bx_{gr}$ in $\xg$). 
Now we successfully seperate the generated space $\pg$ into two parts. To classify $x_{gr}$ and $x_{gf}$ from $\bx_g$ with $D$ as a binary classifier learned the distribution of $\bx_g$ from $p$, we need to minimize its expected miss-classification rate $R(D)$. The loss function for minimizing $R(D)$ by a given $\pi$ could be:
\begin{equation} \label{pu_base}
\begin{split}
\underset{D}{\min} R(D) &= \pi \mathbb{E}_{x \sim \pgr (\bx)}[\ell(D(\bx), 1)] \\
 &+ (1-\pi)\mathbb{E}_{\bx \sim \pgf (\bx)}[\ell(D(\bx), -1)],
\end{split}
\end{equation}
where $\ell(D(\bx), t)$ is the loss function measuring the loss of prediction $D(x)$ when the ground true label is $t$. However, we has less idea about which is $\bx_{gr}$. In our definition, the good samples are similar to the real data, which means that $\pgr$ can be replaced by $\pd$. Thus, the $\pg (\bx)$ can be calculated by:
\begin{equation} \label{u marginal density}
    \pg(\bx) = \pi \pd (\bx) + (1-\pi) \pgf (\bx).
\end{equation}
Similarity, the bad generated samples $x_{gf}$ are also unknown, and we can only access the generated samples $\xg$ and the real data $\xd$. $R(D)$ should be modified to avoid the term of $p_{gf}$. From \eref{u marginal density}, the low-quality part $\pgf$ can be expressed as follows,
\begin{equation} \label{pu_pgf}
\begin{split}
(1-\pi) p_{gf}(\bx) = p_g(\bx) - \pi p_{data}(\bx).
\end{split}
\end{equation}
Then we can find out the follow equation:
\begin{equation} \label{replacing negative}
\begin{split}
	(1-\pi)\mathbb{E}_{\pgf}[\ell(D(\bx), -1)] &= \mathbb{E}_{\pg}[\ell(D(\bx), -1)] \\
& - \pi \mathbb{E}_{\pd}[\ell(D(\bx), -1)].
\end{split}
\end{equation}
By combining Eqs.~(\ref{pu_base}) and (\ref{replacing negative}), the new objective function will be:
{\small\begin{equation} \label{pu based}
\begin{split}
    \underset{D}{\min}~ R(D)& = \pi \mathbb{E}_{\pd}[\ell(D(\bx), 1)] \\
   & + \mathbb{E}_{\pg}[\ell(D(\bx), -1)]
    - \pi \mathbb{E}_{\pd}[\ell(D(\bx), -1)].
\end{split}
\end{equation}
}
\noindent

\renewcommand{\algorithmicrequire}{\textbf{Input:}}
\renewcommand{\algorithmicensure}{\textbf{Output:}}
\begin{algorithm}[t]
\caption{Implmeneting PU learning in GAN}
\begin{algorithmic} 
\REQUIRE The number of $D$ iterations pre $G$ iteration $n_d$ ($n_d = 1$ in normal), the batch size $m$, the class prior knowledge $\pi$ for the proportion of positive data in unlabeled data.
\REQUIRE Initialize the parameters $\theta_g$ of the generator $G$ and the parameters $\theta_d$ of the discriminator $D$.
\WHILE {$\theta_g$ has not converged}
\FOR{$t = 1, ...,n_d$}
\STATE {Randomly sample $\{\bx^{i}\}_{i=1}^{m}\sim \pd(\bx)$;}
\STATE {Randomly sample $\{\bz^{i}\}_{i=1}^{m}\sim \pz(\bz)$;}
\STATE {Sample $\{\bz^{(i)}\}_{i=1}^{m}\sim \mathbb{P}_z$}
\STATE {Calculate $\widehat{R}_{p}^{+} \leftarrow \frac{1}{m}\sum_{i=1}^{m}[f_1(D(\bx^{i}))]$}
\STATE {Calculate $\widehat{R}_{p}^{-} \leftarrow \frac{1}{m} \sum_{i=1}^{m}[f_2(D(\bx^{i}))]$}
\STATE {Calculate $\widehat{R}_{u}^{-} \leftarrow \frac{1}{m} \sum_{i=1}^{m}[f_2(D(G(\bz^{i})))]$}
\STATE {Update $\theta_d \leftarrow \nabla_{\theta_d} \pi\widehat{R}_{p}^{+} + max(0, \widehat{R}_{u}^{-} - \pi\widehat{R}_{p}^{-})$ } 
\ENDFOR

\STATE {Randomly sample $\{z^{i}\}_{i=1}^{m}\sim \pz(z)$;}
\STATE {Update $\theta_g \leftarrow -\nabla_{\theta_g} \frac{1}{m} \sum_{i=1}^{m}[f_2(D(G(z^{i})))]$}
\ENDWHILE
\ENSURE{A generator network $G$.}
\end{algorithmic}
\end{algorithm}
\noindent
By minimizing \eref{pu based}, the discriminator $D$ can distinguish not only $\xgr$ but also $\xd$ from $\xgf$, by only learning the distribution of $\xd$ based on $\pd$ and distribution of $\xg$ (the generated samples) from $\pg$. We notice that the second and third terms of \eref{pu based} are introduced from \eref{pu_pgf} and aim to calculate the loss over $\pgf$. The original loss $(1-\pi)\mathbb{E}_{\pgf}[\ell(D(\bx), -1)]$ is expected to be not less than zero, but the replacement loss function $\mathbb{E}_{\pg}[\ell(D(\bx), -1)] - \pi \mathbb{E}_{\pd}[\ell(D(\bx), -1)]$, may be negative. This abnormal value of loss may lead to over-fitting. It is important to avoid the it to be negative. Finally, the objective function of discriminator proposed in \eref{pu based} will be:
\begin{equation}\label{eq_sd}
\begin{split}
	\underset{D}{\max}~ V(D) & = \pi \mathbb{E}_{\pd}[\log (D(\bx))] \\
    & + \max\{0, \mathbb{E}_{\pz}[\log(1-D(G(\bz)))] \\ 
    & - \pi \mathbb{E}_{\pd}[\log(1-D(\bx))]\}.
\end{split}
\end{equation}
Here we obtain the proposed objective function of the discriminator. Considering there is also an adversarial game between the discriminator $D$ and the generator $G$, the generator $G$ should still be trained to deceive the discriminator $D$. As a result, we can easily lead to the objective function of $G$ as follows,
\begin{equation}\label{eq_sg}
\begin{split}
	\underset{G}{\min}~ V(G) & = \mathbb{E}_{\pz}[\log(1-D(G(\bz)))].
\end{split}
\end{equation}
Following Eqs. (\ref{eq_sd}) and (\ref{eq_sg}), we reached our objective to deal with the generated data in different ways, rather than treating it all as negative samples. Although we finally got a new loss function, in the next section, we theoretically prove that our proposed algorithm is also designed to minimize the distance between the generated distribution and the real distribution, which provides a theory for the effectiveness of our algorithm.

\subsection{PU classification for general GANs}
Above we conclude our objective function within the standard GAN framework. The proposed method can also be integrated into other general GAN frameworks flexibly. In this part, we combine the proposed method with other loss functions of discriminator in GAN. 

In general, the objective function Eq. (\ref{general_d}) contains two loss functions $f_1$ and $f_2$. Those concrete loss functions can be changed for a different variance of GANs, but all these loss functions are following the same concepts that \(f_1(\bx)\) and \(f_2(\bx)\) are trying to separate the real data from the generated data as far as possible. Similar to SGAN, we propose that the discriminator in GAN is better to focus on generated samples with low quality and recognize the high-quality samples from the generated samples. Following this concept, we implement the proposed method for the general framework of GAN with the following equation:
\begin{equation}
\begin{split}
	\underset{G}{\max}~\underset{D}{\min}~ V(D, G) & = \pi \mathbb{E}_{\pd}[f_{1} (D(\bx))] \\
	&+ \max \big\{ 0, \mathbb{E}_{\pz}[f_{2}(D(G(\bz)))] \\
	&- \pi \mathbb{E}_{\pd}[f_{2}(D(\bx)] \big\}.
\end{split}
\label{eq_puall}
\end{equation}
With the help of \eref{eq_puall}, we can now integrate the proposed method into various frameworks of GAN, such as WGAN-GP~\cite{salimans2016improved}, LSGAN~\cite{mao2017least}, and SpectualGAN~\cite{miyato2018spectral}. This flexibility that combining with other models provides the proposed method a chance to get further improvement on existing excellent models. Loss functions corresponding to the specific model can be found in the supplementary material.


%
%
%

\section{Theoretical Analysis}\label{theoretical_results}
In the proposed method, the discriminator $D$ is encouraged to not only distinguish the real samples from the generated samples but also allow a certain proportion of generated samples of high quality to be recognized as real data, which reduces the instability problem during the training progress. Following this principle, we have obtained a novel loss function \eref{eq_sd} and \eref{eq_sg} in the framework of the standard GAN. Although \eref{eq_sd} and \eref{eq_sg} are designed to achieve the above-mentioned desirable characteristics, it is unclear whether the final convergence of the proposed method satisfies the requirements of the generation task. In this section, we provide a formal technical analysis of the convergence of the proposed objective function and prove that the proposed algorithm will perfectly lead the generated distribution to the real one. See the supplementary material for proof.

Now we consider the standard GAN based framework and analyze the optimal discriminator and generator. The discriminator $D$ is optimized by \eref{eq_sd}. Following the analysis proposed in GAN~\cite{goodfellow2014generative}, the optimal distribution $D$ will balance between the true distribution $\pd$ and the learned distribution $\pgf$.
\begin{theorem}\label{thm_1}
For the generator $G$ fixed, the optimal discriminator $D$ is
\begin{equation}\nonumber
\begin{split}
    D^{*}(\bx) =\frac{\pd (\bx)}{\pd (\bx) + \frac{1-\pi}{\pi}\pgf (\bx)},
\end{split}
\end{equation}
where $p_gf(\bx)$ is the distribution of low-quality generated samples of $G$.
\end{theorem}\noindent
With the optimal discriminator $D$ fixed, we can reformulate the objective function by replacing $D(\bx)$ in \eref{eq_sg} according to Theorem~\ref{thm_1}. By doing so, we can summarize the behavior of $G$ in the following theorem.
\begin{theorem}
With the optimal discriminator $D$ fixed, the optimization of generator $G$ is equivalent to minimize $\pi\log\pi + \pi KL(\pd || \pg) + (1-\pi)KL(\pgf || \pg)$.
\label{thm_2}
\end{theorem}\noindent
Theorem~\ref{thm_2} suggests that the optimal generator $G$ will pay attention to reducing the divergence within the generated space by minimizing the distance between $\pg$ and $\pgf$. Moreover, the generator will also guide the generated distribution $\pg$ as close as possible to the real distribution $\pd$, which ensures the quality of the generated sample. Combining Theorem~\ref{thm_1} and Theorem~\ref{thm_2}, we can summarize the following corollary.
\begin{figure*}[t]
\begin{center}
\includegraphics[width=0.85\linewidth]{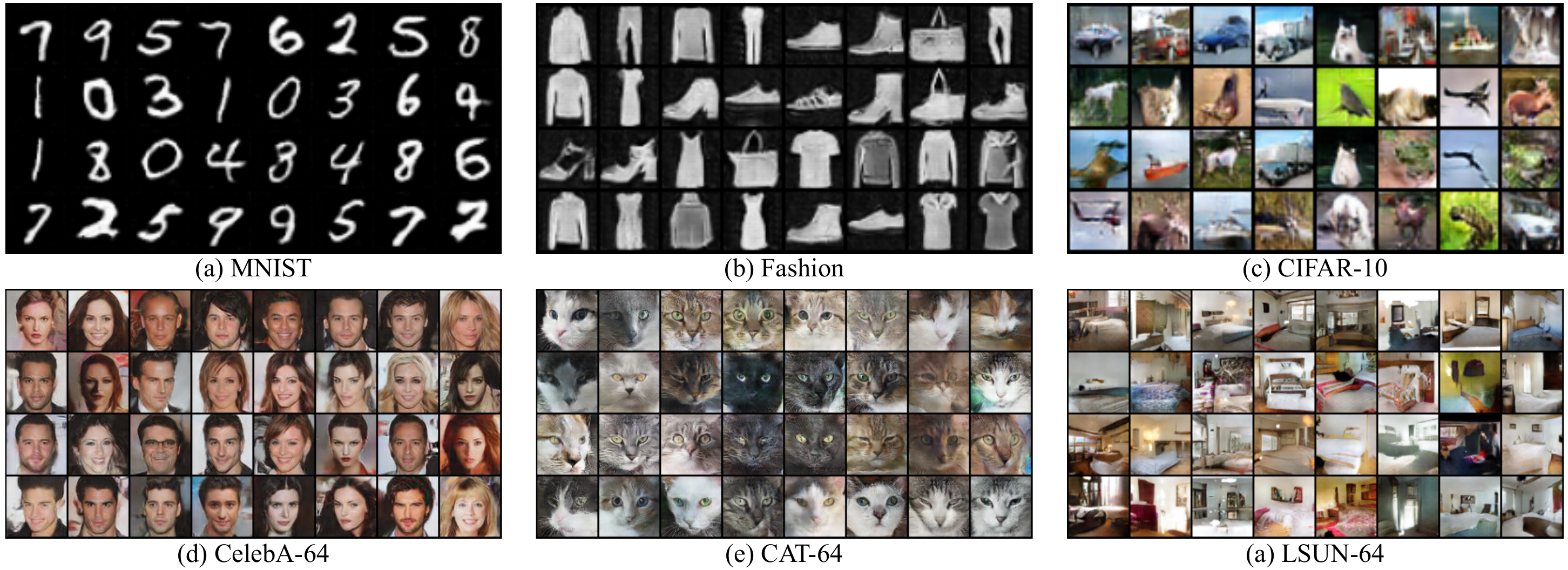}
\end{center}
\vskip -0.25in
   \caption{Generated samples obtained by the proposed method on image datasets.}
\label{fig_gen}
\vspace{-0.15in}
\end{figure*}

\begin{table*}[ht]
\centering
\small
  \caption{Performance comparison on several benchmark datasets.}
  \vspace{-0.1in}
  \begin{tabular}{c|cc|cc|cc|cc}
    \toprule
    Loss     &  SGAN &  PUSGAN  &  HingeGAN &  PUHingeGAN  &  LSGAN &  PULSGAN  &  WGAN-GP &  PUWGAN-GP\\
    \midrule
    MNIST & 18.65 & 16.53 & 21.48 & 16.91 & \textbf{13.47} & 13.96 & 17.42 & 14.77 \\
    Fashion & 25.72 & 24.33 & 28.39 & 25.31 & 32.05 & 26.72 & 26.50 & \textbf{23.12} \\
    CIFAR-10 & 43.39 & 31.02 & 43.85 & 36.37 & 27.64 & \textbf{22.32} & 36.86 & 31.85 \\
    CelebA-64 & 48.44 & 43.89 & 46.13 & 43.94 & 51.67 & 47.80 & 36.09 & \textbf{35.72} \\
    CAT-64 & 46.13 & \textbf{16.90} & 29.52 & 25.72 & 57.22 & 26.79 & 21.86 & 17.35 \\
    \bottomrule
  \end{tabular}
\label{tab_big}
\vspace{-0.25in}
\end{table*}
\begin{corollary}
The global minimum of the proposed objective function $V(G, D)$ is achieved if and only if $\pgf = \pg = \pd$. At that point, $C(G)$ achieves the value of $\pi\log\pi$, and $D(\bx)$ achieves the value of $\pi$.
\label{coro_1}
\end{corollary}\noindent
The above theoretical results prove that the proposed method will achieve equilibrium if and only if $\pg=\pd$, which points out that our method enjoys the same global equilibrium point as other GAN frameworks. These results justify our approach. The next experiment section further illustrates the effectiveness of our approach.

\section{Experiment}
In this section, we evaluate the proposed method on a range of datasets including MNIST~\cite{lecun1998gradient}, FMNIST~\cite{xiao2017/online}, CIFAR-10~\cite{krizhevsky2009learning}, CAT~\cite{zhang2008cat}, and LSUN-bedroom~\cite{yu15lsun}. We resize images in the MNIST and FMNIST datasets to $32\times32$ for convenience. For these datasets with more than one kind of resolution, we mark them with the resolution, such as CAT-64. Experiments on CAT-128, CAT-256, CelebA-128, and LSUN-128 datasets are also conducted to evaluate the high-resolution generation ability of the proposed approach. Moreover, due to the limitation of computational resource, for LUSN-128 we randomly sample 100,000 images from the dataset as training set, instead of using all of them. The experiment is implemented in pytorch, and we use FID (Fréchet Inception Distance) as the quantitative indicator to evaluate the performance based the quality of the generated results (lower value of FID indicating higher generated quality). FID scores are calculated with 10,000 generated samples and 10,000 real images randomly sampled from the dataset in advance.


As mentioned in Section~\ref{sec_3}, our approach enjoys a high degree of flexibility can be integrated into most kind of GAN frameworks. We chose some variants of GAN as basic frameworks, such as standard GAN (SGAN)~\cite{goodfellow2014generative}, LSGAN~\cite{mao2017least}, WGAN-GP~\cite{gulrajani2017improved}, and HingeGAN~\cite{miyato2018spectral}, and then integrate our method into these frameworks for comparison. For a fair comparison, we follow the same settings and architectures of GANs when we integrate our method and make sure the loss functions are the only changed part. We also compare our method with Relativistic GAN~\cite{jolicoeur2018relativistic}, which is another flexible GAN framework, and we use the average version (RaGAN) in the experiment. All objective function of the proposed frameworks could be found in the supplementary material.


 All models used in experiments will be trained with Adam optimizer~\cite{kingma2014adam}, and the random selecting seed is set to 1. In addition, the discriminator follows the CNN structure described by Miyato, et al. (2018)~\cite{miyato2018spectral} while the generator will follow the structure of standard DCGAN~\cite{radford2015unsupervised} for all models generating images whose resolution less than 128 except WGAN-GP whose structure we leave in the Supplementary material. We use the stable setting for DCGAN~\cite{radford2015unsupervised} as the \textit{basic setting} for training, which the learning rate $lr$ is set to 0.0002, the \(\beta_1 = 0.5\) and \(\beta_2 = 0.999\) for Adam optimizer, and the number of training time for discriminator and generator will both equal to 1. In addition, batch normalization~\cite{ioffe2015batch} is also implemented. Moreover, we set a general case for the hyper-parameter $\pi$ growth pattern called the \textit{basic pattern} in this experiment. The basic pattern will be initialized $\pi$ with 0.1 and will increase smoothly at each iteration until it reaches 0.7. Detailed network structures used on other datasets could be found in the supplementary material.

\begin{table*}[ht]
\caption{Stability experiment results (reported by FID) on the CAT dataset with three resolution.}
\vspace{-0.1in}
  \centering
  \small
    \begin{tabular}{c|cccc|cccc|cccc}
    \toprule
    \multirow{2}*{Loss} &  Min & Max & Mean & SD &  Min & Max & Mean & SD &  Min & Max & Mean & SD\\
    &\multicolumn{4}{c|}{64$\times$64 images (N=9304)}&\multicolumn{4}{c|}{128$\times$128 images (N=6645)} & \multicolumn{4}{c}{256$\times$256 images (N=2011)}  \\
    \midrule
    SGAN & 19.32 & 95.31 & 48.36 & 33.92 & - & - & - & -  & - & - & - & -    \\
    RaSGAN     & 19.11 & 42.20 & 32.39 & 7.53 & 23.18 & 38.74 & 26.12 & 5.66 & 33.67 & 133.94 & 60.42 & 23.66    \\
    PUSGAN     & \textbf{12.71} & \textbf{18.99} & \textbf{16.38} & 1.99 & 17.76 & \textbf{28.91} & 23.61 & \textbf{3.43} & 35.75 & \textbf{63.68} &\textbf{50.76} & \textbf{9.9}    \\
    \midrule
    LSGAN & 21.84  & 69.03 & 39.22 & 13.64 & 22.75  & 62.39 & 43.48 & 12.68 & - & - & - & -    \\
    RaLSGAN & 15.73 & 25.20 & 19.85 & 4.47 & \textbf{16.52} & 43.87 & \textbf{23.74}  & 7.33 & 42.02  & 282.15 & 72.32 & 86.11   \\
    PULSGAN & 12.86 & 26.79 & 21.28 & 4.36 & 18.27 & 38.52 & 26.24  & 6.06 & \textbf{33.23}  & 269.27 & 73.68 & 89.04   \\
    \midrule
    WGAN-GP & 17.73  & 31.62 & 24.61 & 5.19 & 21.91 & 43.06 & 34.48 & 8.24 & 58.31 & 262.24 & 109.21 & 64.38     \\
    RaSGAN-GP & 16.83  & 24.32 & 20.59 & 2.84 & 18.78 & 44.53 & 31.63 & 10.04 & 39.42 & 115.07 & 69.65 & 24.19     \\
    PUWGAN-GP & 13.60  & 20.67 & 17.13 & \textbf{1.73} & 17.32 & 36.17 & 27.32 & 5.52 & 35.29 & 93.48 & 59.01 & 17.71     \\
    \bottomrule
    \end{tabular}
\label{tab_cat}
\vspace{-0.2in}
\end{table*}

\subsection{Quantitative Image Generation Results}
In this section, we evaluate the generation ability of the proposed method on multi-category image sets MNIST~\cite{lecun1998gradient}, FMNIST~\cite{xiao2017/online}, and CIFAR-10~\cite{krizhevsky2009learning} and single-category image sets CAT~\cite{zhang2008cat} and LSUN-bedroom~\cite{yu15lsun}. In this part, the resolution of images in the MNIST, FMNIST, and CIFAR-10 is 32 and is 64 in the rest datasets. We choose four representative adversarial models SGAN, HingeGAN, LSGAN, and WGAN-GP as basic frameworks and compare them with our algorithms, which are denoted as PUSGAN, PUHingeGAN, PULSGAN, and PUWGAN-GP, respectively.

Table~\ref{tab_big} reports a comparison of the FID score obtained by the proposed method and basic models. Our models enjoy the ability to combine with most variants of GAN, which allows us to achieve the best performance on most data sets. Table~\ref{tab_big} shows that most of our methods exceed their corresponding basic frameworks, which demonstrates the effectiveness and the flexibility of our approach. In Figure~\ref{fig_gen}, we show a few images generated by the PUSGAN models. We observe that the proposed method generates high-quality images on various datasets, which is consistent with the quantitative results in Table~\ref{tab_big}.

\begin{table}[t]
\centering
  \caption{Performance comparison based on FID on the CIFAR-10 dataset with different settings.}
  \vspace{-0.1in}
  \begin{tabular}{c|cccc}
    \toprule
    Loss     &  basic &  $lr=.001$  & No BN & Tanh \\
    \midrule
    SGAN & 43.39 & 74.75   & 47.83 & 57.42   \\
    RaSGAN     & 33.63 & 44.67  & 42.21 & 55.38  \\ 
    PUSGAN & 31.02 & 40.93   & 37.85 & 54.70  \\
    \midrule
    LSGAN & 27.64  & 56.36   & 40.81 & 68.40 \\ 
    RaLSGAN     & 23.48 & \textbf{35.75}  & \textbf{37.28} & 55.92 \\
    PULSGAN     & \textbf{22.32}  & 37.65   & 37.34 & 51.45  \\
    \midrule
    HingeGAN & 43.85  & 41.66   & 39.57 & 58.43\\
    RaHingeGAN     & 38.03 & 44.25  & 41.54 & 51.70 \\   
    PUHingeGAN & 36.37  & 35.60  & 38.09 & \textbf{50.84}  \\
    
    \bottomrule
  \end{tabular}
  \label{tab_cirfar10}
  \vspace{-0.25in}
\end{table}

\subsection{Evaluating Stability}
We evaluate the stability of the proposed method on three resolution of the CAT dataset, such as 64 $\times$ 64, 128 $\times$ 128, and 256 $\times$ 256 pixels. As there are only 6654 and 2011 samples in the CAT-128 dataset and CAT-256 dataset respectively, some variants of GAN are unable to converge on these datasets. We choose SGAN, LSGAN, and WGAN-GP as basic models. We compare the proposed method with both these basic models as well as the corresponding Relativistic GAN (RaGAN). For each model, We calculate the FID score of the current model every 10,000 iterations. The results will be presented with the minimum, maximum, mean, and standard deviation (SD) of these obtained FID values. Table \ref{tab_cat} shows the FID results for different networks in different resolutions of data sets. 

For 64$\times$64 resolution dataset, all models trained by the proposed method except PULSGAN, can achieve much lower FID in minimum, maximum and mean compared with its original version and even can further achieve lower FID values than their relativistic versions, which indicates that our algorithm can effectively improve the training stability and improve the quality of the generation.

For higher resolution data sets, SGAN failed to converge in 128x128 and 256x256 resolution datasets while LSGAN will be stuck at the early stage in 256x256 resolution dataset~\cite{jolicoeur2018relativistic}. The standard version of our model (PUSGAN) shows further stability with lower values of maximum, mean, and SD in all three resolution datasets. On the most challenging 256 resolution dataset, the proposed method achieves both the satisfactory quality and stability. In experiments, we found that although the PULSGAN can converge in the CAT-256 dataset, the convergence is much slower than other GANs. Nevertheless, PULSGAN still can achieve competitive results compared with other states of arts GANs such as WGAN-GP and RaGAN.

Overall, our algorithm presents desirable stability for all three data sets, and it can achieve similar or even better results compare to relativistic versions. It is impressive that all these GANs trained by the proposed method have improved. As a result, we conclude that the above stability experiment demonstrates that the proposed method provides could provide stability for the training progress for a variety of GANs and thus improve the quality of generated images.


%
%

\begin{figure*}[t]
\begin{center}
\includegraphics[width=0.75\linewidth]{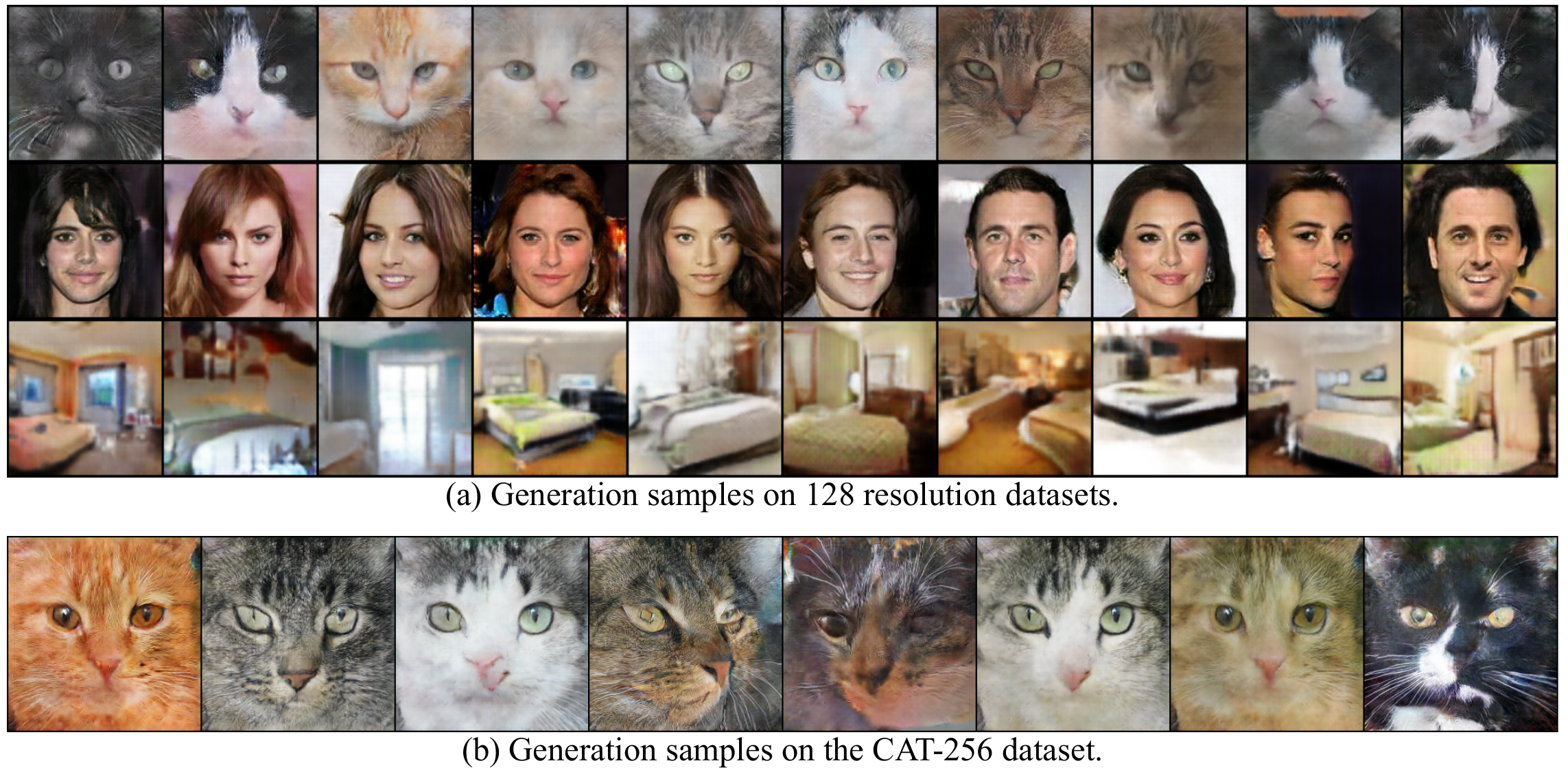}
\end{center}
\vskip -0.25in
   \caption{High-resolution generated samples.}
\label{fig_gen_hq}
\vspace{-0.1in}
\end{figure*}

\begin{figure*}[t]
\begin{center}
\includegraphics[width=0.75\linewidth]{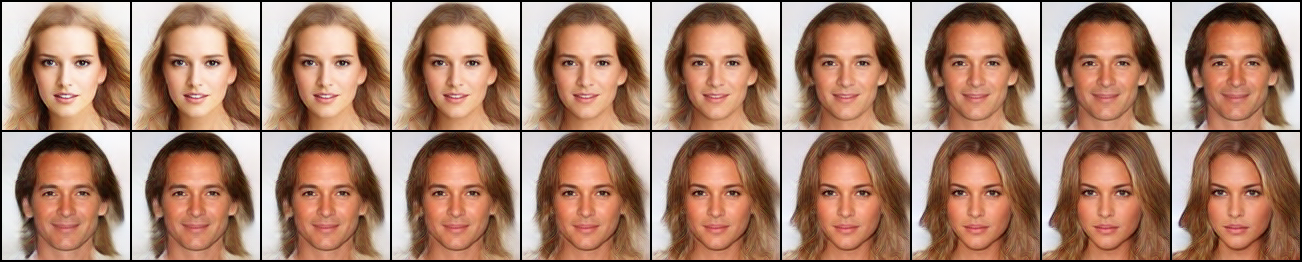}
\end{center}
\vskip -0.25in
   \caption{High-resolution generated samples.}
\label{fig_int_hq}
\vspace{-0.2in}
\end{figure*}

\subsection{Evaluating on Hard Training Setting}
As we have claimed, our approach focuses on improving low-quality samples and lead to more stable training progress. Thus our approach enjoys the ability to generalize to many training settings. To demonstrate this, we evaluate the proposed method on several hard training settings and compare it with the other GAN frameworks. In this part, we implement SGAN (RaSGAN, PUSGAN), LSGAN (RaLSGAN, PULSGAN), and HingeGAN (RaHingeGAN, PUHingeGAN) on the CIFAR-10 dataset. The experiment is conducted on one basic setting and three hard settings. The basic setting is same as above, and three hard settings are i) changing learning rate to $0.001$ (lr=.001), ii) removing Batch normalization layers in $G$ and $D$ (No BN), and iii) replacing all activation functions with Tanh in $G$ and $D$ (Tanh).

The results are showed in Table \ref{tab_cirfar10}. In the stable setting, we can find that the PUSGAN has better performance than SGAN and its other variants. The PUHingeGAN has a huge improvement compared with the original HingeGAN with a gap of 14, and it also performs better than RaHingeGAN. On the other hands, the performance of PULSGAN is slightly worse than other LSGAN versions.


When the learning rate is increased to 0.001, all three PU versions of GANs perform well, compared with the original one. While PUSGAN and PUHingeGAN can perform better than their relativistic versions. However, by changing optimization settings such as removing batch normalization or replace ReLU activation function with Tanh activation function (No BN and Tanh in columns respectively), the performances of PUGANs will be worse. It might indicate that the PUGANs rely on optimization terms for stable training.

\subsection{High-resolution Results}
\vspace{-0.05in}
Generating high-resolution images is a complicated task. To demonstrate the generation ability of our algorithm, we evaluate the proposed method on CAT-128, CelebA-128, and LSUN-128 datasets with 128 $\times$ 128 pixels and the CAT-256 dataset with 256 $\times$ 256 pixels. There are 202,599 images in the CelebA-128 dataset, and 3,033,042 in the LSUN-128 dataset (only 100,000 samples are used for training). As mentioned above, the CAT-128 and CAT-256 datasets are considered as more challenging high-resolution datasets because there are only 6,645 and 2,011 samples, respectively. Images shown in Figure~\ref{fig_gen_hq} are generated by the proposed method within the architecture of SGAN (PUSGAN), while SGAN failed to generate such high-resolution images, especially on the CAT-256 dataset. Moreover, interpolation is also an impressive feature of the generative models, which indicates that the generative model successfully learns to fit the distribution of natural images instead of overfitting to the training samples. We show high-resolution interpolation results obtained by the proposed method in Figure~\ref{fig_int_hq}. It shows that our model generates smooth interpolation images. Figures~\ref{fig_gen_hq} and ~\ref{fig_int_hq} demonstrate that the proposed method could provide improvement on the quality of generated samples.


\begin{figure}[t]
\begin{center}
\includegraphics[width=.85\linewidth]{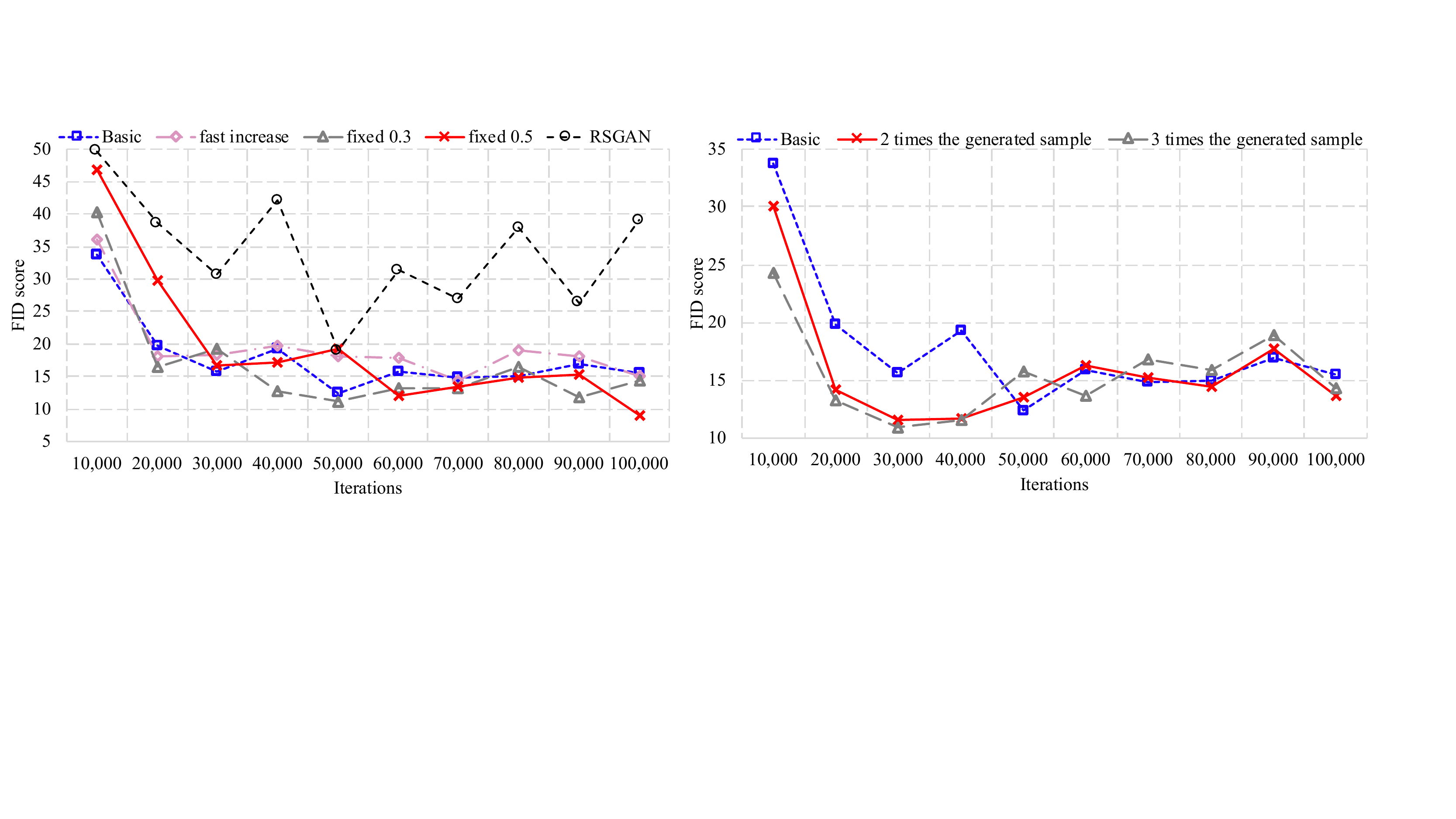}
\end{center}
\vspace {-0.2in}
   \caption{The trend of Fréchet Inception Distance (FID) of PUSGAN using different $\pi$ growth pattern.}
\label{fig:pi}
\vspace{-0.3in}
\end{figure}

\subsection{Evaluating the impact of class prior $\pi$}
\vspace{-0.05in}
In \eref{eq_pi}, we introduce a class prior $\pi$ into our algorithm. The $\pi$ indicates the proportion of high-quality fake data in fake data, and we treat it as a hyper-parameter. In this section, we further evaluate the impact of class prior $\pi$ with PUSGAN framework on the CAT-64 dataset. The structure and training settings are the same as the previous sections. We set four different increasing patterns for $\pi$ during training. The first pattern is the basic pattern we used in previous sections. The second version will set $\pi$ to be 0.3 at the beginning, and it will be increased with 0.1 at every 10k iterations until 0.7 is reached. The second version is used to evaluate the 
impact of the fast growth of $\pi$. For the third and fourth patterns,
$\pi$ will be fixed at 0.3 and 0.5 during the training process. 

    





In Figure~\ref{fig:pi}, we found that the fast-growing pattern achieves the worst average performance, and its FID scores remain relatively high. As a comparison, the basic pattern can reach lower FID values than the fast pattern, and it is also relatively stable in the later training stage. The fixed $\pi$ value of 0.3 could present more stability and generate a competitive result, compared with the previous two patterns. In addition, the fourth version has the worst performance at the beginning, but it was keep going better and achieved the best performance in the end, within all four patterns. The shortage of this version mainly lies in the large fluctuations and slow convergence. It is interesting that all first three patterns have similar FID values at the early stage, while the performance with a higher $\pi$ is much worse at the same stage. This may be because the proportion of high-quality samples in the $G$ network at the beginning of training is far from 0.5, and setting $\pi$ to 0.5 is against the real situation, leading to an unstable training. On the other hands, the fourth one achieves a lower FID value in the end, while others have similar values at the same time. It might show a too large (\eg 0.7), or a too small (\eg 0.3) values of $\pi$ reduce the performances. The result shows that the $\pi$ can affect the performance of the generation. We also present the result obtained by RaGAN for comparison. It shows that all the four versions of the model obtained by our algorithm produce both higher quality images and show better stability. As a result, the proposed method enjoys considerable tolerance for the selection of hyperparameters.

\begin{figure}[t]
\begin{center}
\includegraphics[width=0.85\linewidth]{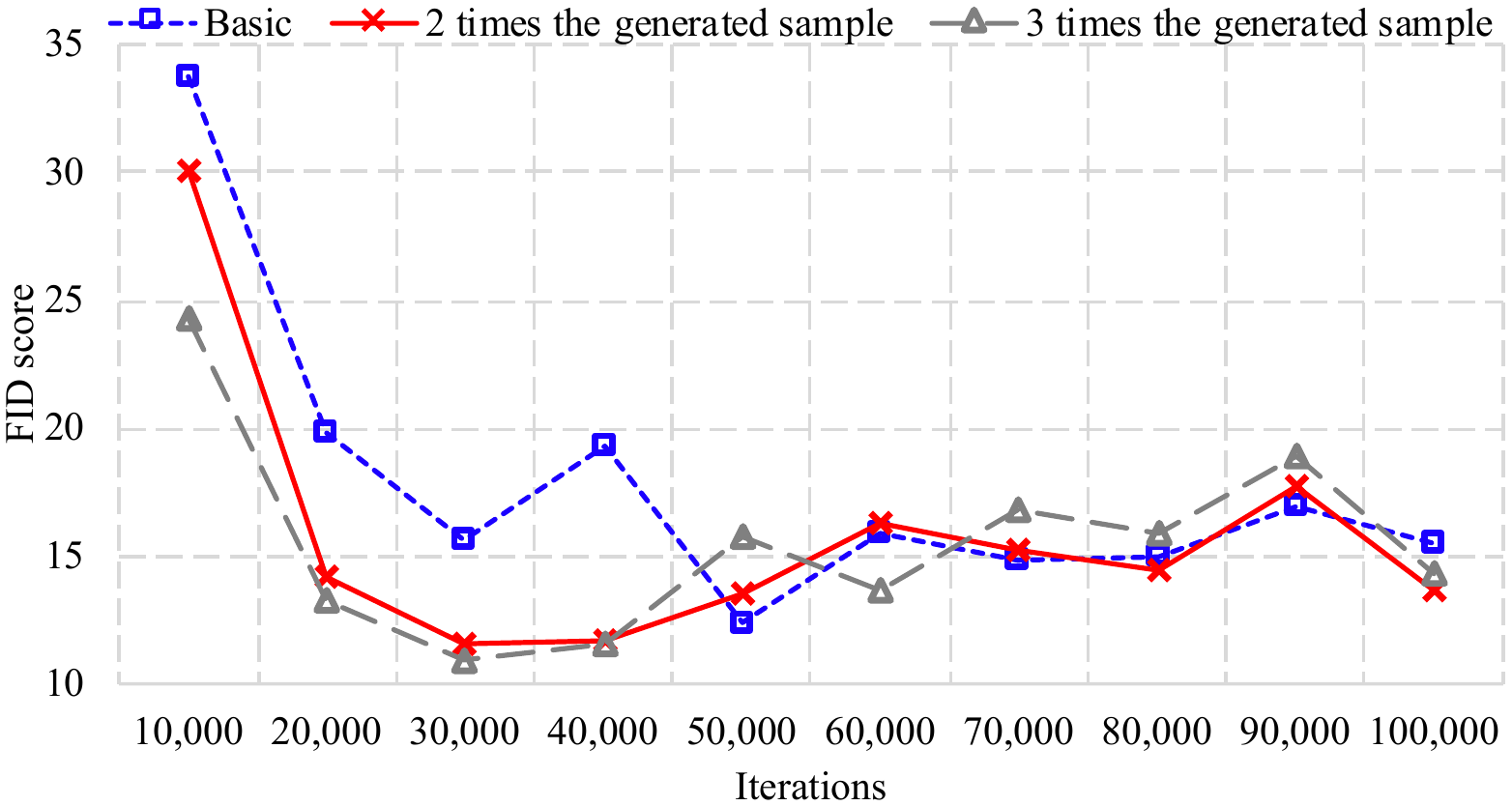}
\end{center}
 \vskip -0.2in
   \caption{The trend of Fréchet Inception Distance (FID) of PUSGAN trained by different batch sizes of fake data.}
\label{fig:batch_size}
 \vspace{-0.25in}
\end{figure}


\subsection{Evaluating the impact of increasing number of fake data for training}
\vspace{-0.05in}
Normally, the size of real data is much smaller than the size of the generated data in the adversarial generative task. It is an interesting problem about how to make the most of this large amount of generated data. In general training progress of GAN, the batch size of real samples and generated samples are the same. Here, we try to increase the batch size of the generated data and maintain that of the actual data to take advantage of this large number of generated samples. We investigate the impact of increasing the batch size of the generated data for training. The evaluation is based on three versions of PUSGAN with different batch sizes of fake data. The first version is the basic version that the batch size of real and fake data are the same. The second and third versions will use twice and three times more fake data than real data, respectively. The structure and training settings are the same as the one we used in previous sections. 
We report these interesting results Figure~\ref{fig:batch_size}.

From results, we find that the second and third versions of PUSGAN can reach their best performance at the very beginning, which proves that PUSGAN can converge faster by increasing the batch size of fake data. Although it shows that an increase in the number of false samples can provide a small performance boost and faster convergence, it seems to be detrimental to stability. Considering the stability and for a fair comparison, we insist on using the same batch size for both the real and generated data in the above experiments.

\section{Conclusion}
In this paper, we present a positive-unlabeled generative adversarial network (PUGAN), where the discriminator is trained to recognize the high-quality samples from the generated data, to obtain a more stable training progress. The proposed method addresses problems in traditional methods that neglecting the gradual increase in sample quality and the imbalance of generated sample quality, which provides more stable training progress and higher generation quality.
We further demonstrate that our approach has the flexibility to combine with most existing GAN frameworks without requiring the addition of computational cost. Experiments conducted on real-world image datasets suggest that the proposed method successfully improve both the stability and the quality of generated samples. We also provide some theoretical results to illustrate the justification of our approach.


\clearpage

{\small
\bibliographystyle{ieee_fullname}
\bibliography{cvpr20pugan}
}

\end{document}